\documentclass{article}

\usepackage[preprint,nonatbib]{arxiv}

\usepackage[utf8]{inputenc} 
\usepackage[T1]{fontenc}    
\usepackage{hyperref}       
\usepackage{url}            
\usepackage{booktabs}       
\usepackage{amsfonts}       
\usepackage{nicefrac}       
\usepackage{microtype}      
\usepackage{lipsum}

\title{Ensemble Learning Applied to Classify GPS Trajectories of Birds into Male or Female}

\author{
  Dewan Fayzur \\
  DevScope S.A.\\
  Porto, Portugal \\
  \texttt{fayzur20@gmail.com} \\
}

\begin{document}
\maketitle

\begin{abstract}
We describe our first-place solution to the Animal Behavior Challenge (ABC 2018) on predicting gender of bird from its GPS trajectory. The task consisted in predicting the gender of shearwater based on how they navigate themselves across a big ocean. The trajectories are collected from GPS loggers attached on shearwaters' body, and represented as a variable-length sequence of GPS points (latitude and longitude), and associated meta-information, such as the sun azimuth, the sun elevation, the daytime, the elapsed time on each GPS location after starting the trip, the local time (date is trimmed), and the indicator of the day starting the from the trip. We used ensemble of several variants of Gradient Boosting Classifier along with Gaussian Process Classifier and Support Vector Classifier after extensive feature engineering and we ranked first out of 74 registered teams. The variants of Gradient Boosting Classifier we tried are CatBoost (Developed by Yandex), LightGBM (Developed by Microsoft), XGBoost (Developed by Distributed Machine Learning Community). Our approach could easily be adapted to other applications in which the goal is to predict a classification output from a variable-length sequence.
\end{abstract}


\newpage

\section{Introduction}
The Animal Behavior Challenge was organized by the 2018 Symposium on Systems Science of Bio-Navigation \footnote{\url{http://navi-science.org/2017/09/19/symposium-on-systems-science-of-bio-navigation-2018/}}, sponsored by Technosmart \footnote{\url{http://www.technosmart.eu/index.php}} and proposed as a CodaLab competition \footnote{\url{https://competitions.codalab.org/competitions/16283}}. It consisted in classifying the gender of shearwater based on trajectories (latitude and longitude) and some meta-information associated to each shearwaters' trip. Such prediction models could help to understand shearwater more efficiently and how they navigate themselves, like male and female shearwater could use different trajectories along the way of trip.

The training dataset is composed of all the GPS trajectories of 631 streaked shearwaters (326 male and 305 female) breeding on Awashima Island, Japan. Each datapoints in the training dataset representing a complete bird trip and being composed of the following attributes \footnote{\url{https://competitions.codalab.org/competitions/16283\#participate-get_data}}
:

\begin{itemize}
    \item longitude
    \item latitude
    \item sun azimuth: clockwise from the North
    \item sun elevation: upward from the horizon
    \item daytime: 1 being day, or 0 being night
    \item elapsed time: after starting the trip
    \item local time: only time with a format (hh:mm:ss)
    \item days: days after the trip starts
\end{itemize}

In the competition setup, the testing dataset is composed of all the GPS trajectories of 275 streaked shearwaters.
In the Development Phase of the competition 10\% of the submission labels are randomly modified to report score. In the Final Phase of the competition, all the submission are recalculated for the final ranking.

Our approach uses extensive feature engineering prior to ensemble learning. Section \ref{sec:Feature} describes feature engineering techniques. Section \label{sec:TheWinningApproach} introduces our winning model, which is based on a ensemble learning architecture. Section 4 and Section 5 compares and analyses our various models quantitatively and qualitatively on the competition dataset set. The source code of our first-place solution can be found online \url{https://github.com/dfayzur/Animal-Behavior-Challenge-ABC2018}.

\section{Feature Engineering}
\label{sec:Feature}
\subsection{Basic New Features}
\label{sec:BasicFeature}
We first created velocity, acceleration, distance features for each of the GPS points from the gives dataset. At this time we have 7 key features for each GPS points to work with, such features are: velocity, acceleration, distance, longitude, latitude, azimuth, and elevation. We also created the differences of features at time \emph{t} to the next point at time \emph{t+1}. We call these features as \emph{delta} of velocity, longitude, latitude, azimuth, and elevation. From these 12 features, we took quintiles at 0\%, 5\%, 10\%, 20\%, 25\%, 30\%, 40\%, 50\%, 60\%, 70\%, 75\%, 80\%, 90\%, 95\%, and 100\% of them. In addition, we also calculated average, minimum, and maximum of these 12 features. All these operations are applied on each sequence of GPS trajectories for each bird individually. We also calculated the number of times a velocity exceeds to values (average and quintiles at 5\%, 10\%, 15\%, 25\%, 50\%, 75\%, 80\%, 85\%, 90\%, 95\%, and 99\%) calculated over all GPS trajectories combined. Finally we took first 5 longitude and latitude values of each birds' trajectories. We also included Principal component analysis (PCA) on longitude, latitude, azimus, elevation, and velocity of individual birds' and added them to the final features list.

\subsection{Preparing Dataset}
\label{sec:PreparingFeature}
We have created two training dataset based on the features generated from \ref{sec:BasicFeature}. In the first dataset, we splitted the original dataset to day and night trajectories and apply features generation on them according to features described in \ref{sec:BasicFeature}. So, this operation doubles the number of features, and we call this dataset as \emph{split}. In the second dataset, we consider all trajectories together apply features generation process. We call this dataset as \emph{together}. The test dataset were created on similar ways.

\section{The Winning Approach}
\label{sec:TheWinningApproach}
\subsection{Modeling}
\label{sec:Modeling}
We have trained 10 binary predictive models on each of the dataset created in \ref{sec:PreparingFeature}. All together we trained 20 models with 5 fold cross validations, and predicted on the test dataset. The trained models are:
\begin{itemize}
    \item Gradient boosted decision trees \cite{Friedman2001}: Gradient boosting is a machine learning technique for regression and classification problems, which produces a prediction model in the form of an ensemble of weak prediction models, typically decision trees. We modeled variants of Gradient boosted decision trees:
        \begin{itemize}
            \item XGBoost \cite{Chen2016}: a distributed gradient boosting library designed to be highly efficient, flexible and portable. We modeled two version of XGBoost with two different loss functions:
                \begin{enumerate}
                    \item binary logistic: logistic regression for binary classification, output probability. (model name which we call \emph{xgb\_binary})
                    \item pairwise rank: set XGBoost to do ranking task by minimizing the pairwise loss. (model name which we call \emph{xgb\_rank})
                \end{enumerate}
            \item LightGBM \cite{LightGBM2017}: an effective gradient boosting decision tree which contains two novel techniques: Gradient-based One-Side Sampling and Exclusive Feature Bundling to deal with large number of data instances and large number of features respectively. We modeled two version of LightGBM with two different boosting types:
                \begin{enumerate}
                    \item gbdt: traditional Gradient Boosting Decision Tree. (model name which we call \emph{lgb\_gbdt})
                    \item rf: Random Forest. (model name which we call \emph{lgb\_rf})
                \end{enumerate}
            \item CatBoost \cite{Dorogush2017CatBoostG}: a state-of-the-art gradient boosting on decision trees library, which support for both numerical and categorical features. (model name which we call \emph{cat})
            \item GradientBoostingClassifier \cite{Friedman2001}: a Gradient Boosting for classification algorithm from scikit-learn \cite{scikit-learn} library. It builds an additive model in a forward stage-wise fashion and it allows for the optimization of arbitrary differentiable loss functions. (model name which we call \emph{sk\_gbt})
            \item RandomForestClassifier \cite{Breiman2001}: a meta estimator that fits a number of decision tree classifiers on various sub-samples of the dataset to improve the predictive accuracy and control over-fitting. This algorithm is from scikit-learn \cite{scikit-learn} library. (model name which we call \emph{sk\_rf})
            \item ExtraTreesClassifier \cite{Geurts2006}: a meta estimator that fits a number of randomized decision trees on various sub-samples of the dataset to improve the predictive accuracy and control over-fitting. This algorithm is from scikit-learn \cite{scikit-learn} library. (model name which we call \emph{sk\_et})
            \item SVC \cite{Chang2011}: a libsvm based Support Vector Machines estimator from scikit-learn \cite{scikit-learn} library. (model name which we call \emph{sk\_svc})
            \item GPC \cite{Rasmussen2005}: a probabilistic predictions with Gaussian process classification estimator from scikit-learn \cite{scikit-learn} library. (model name which we call \emph{sk\_gpc})
        \end{itemize}    
\end{itemize}

The model parameters are identified by 5 fold cross validations on each of the algorithms on both \emph{together} and \emph{split} dataset. Thus all the model name will be prefixed with \emph{together} and \emph{split} to distinguish. Models estimated on \emph{together} dataset are called: \emph{together\_xgb\_rank}, \emph{together\_xgb\_binary}, \emph{together\_lgb\_gbdt}, \emph{together\_lgb\_rf}, \emph{together\_cat}, \emph{together\_sk\_gbt}, \emph{together\_sk\_rf}, \emph{together\_sk\_et}, \emph{together\_sk\_svc}, and \emph{together\_sk\_gpc}. Models estimated on \emph{split} dataset are called: \emph{split\_xgb\_rank}, \emph{split\_xgb\_binary}, \emph{split\_lgb\_gbdt}, \emph{split\_lgb\_rf}, \emph{split\_cat}, \emph{split\_sk\_gbt}, \emph{split\_sk\_rf}, \emph{split\_sk\_et}, \emph{split\_sk\_svc}, and \emph{split\_sk\_gpc}. The cross validation results are described in the section \ref{sec:Experimental}. Finally, we report the simple majority vote ensemble of all predictions generated from 20 trained models.

\section{Experimental Results}
\label{sec:Experimental}
In the competition setup, wee needed to submit directly target on the classification decision boundary without producing the probability estimation. This kind of prediction is called hard classification problem, in contrast to soft classification problem where we could submit probability estimation. The evaluation matric is Accuracy which is also hard to optimize if the prediction problem is unbalanced. We decided to train all the models listed in \ref{sec:Modeling} to optimize for better F1 score \cite{Goutte2005}. The F1 score is the harmonic average of the precision and recall, often perform better in classification problem settings. Thus, F1 score was the stoping criterion at the time of finding optimal parameters of each models. 

\subsection{Cross Validation}
\label{sec:Cross}
As the competition testing dataset is small and submission of predicted classification is randomly modified and reported, we can not reliably compare models on leaderboard results. Thus we decided stick on the cross validation F1 score to tune models and select models in the development phase. A 5 fold cross validation \cite{Stone1974} strategies was used throughout the development phase. The 5 fold cross validation split is same for all models and for both custom  \emph{together} and \emph{split} dataset. Table ~\ref{tab:table} shows the 5 fold cross validation scores of our various models on our custom dataset as well as on the ensemble ones. The different hyperparameters of each model have been tuned and used in the final submission models in subsection \ref{sec:Submissions}. 

\begin{table}
 \caption{5 fold cross validation F1 score}
  \centering
  \begin{tabular}{lll}
    \toprule
    \multicolumn{2}{c}{Model}                   \\
    \cmidrule(r){1-2}
    Type     & Name            & F1 score (cross validation) \\
    \midrule
    Base     & \emph{together\_xgb\_rank}   & 0.771361  \\
    ~     & \emph{together\_xgb\_binary} & 0.768704  \\
    ~     & \emph{together\_lgb\_gbdt}   & 0.765479  \\
    ~     & \emph{together\_lgb\_rf}     & 0.758018  \\
    ~     & \emph{together\_cat}         & 0.732773  \\

    ~     & \emph{together\_sk\_gbt}     & 0.768177  \\
    ~     & \emph{together\_sk\_rf}      & 0.761317  \\
    ~     & \emph{together\_sk\_et}      & 0.761673  \\
    ~     & \emph{together\_sk\_svc}     & 0.725569  \\
    ~     & \emph{together\_sk\_gpc}     & 0.717157  \\

    ~     & \emph{split\_xgb\_rank}      & 0.767530     \\
    ~     & \emph{split\_xgb\_binary}    & 0.767530      \\
    ~     & \emph{split\_lgb\_gbdt}      & 0.767029  \\
    ~     & \emph{split\_lgb\_rf}        & 0.755920  \\
    ~     & \emph{split\_cat}            & 0.734398  \\

    ~     & \emph{split\_sk\_gbt}        & 0.765193     \\
    ~     & \emph{split\_sk\_rf}         & 0.759317       \\
    ~     & \emph{split\_lgb\_et}        & 0.752338  \\
    ~     & \emph{split\_lgb\_svc}       & 0.725369  \\
    ~     & \emph{split\_lgb\_gpc}       & 0.714567  \\
    \bottomrule
    Ensemble     & Ensemble       & 0.793456  \\
    \bottomrule
  \end{tabular}
  \label{tab:table}
\end{table}

\subsection{Final Submission}
\label{sec:Submissions}
To reduce the effect of random seeds in the 20 models setting, we simulated each model settings for 10 times with 10 different random seeds. For each model setting, we used the hyperparameters found in the previous subsection \ref{sec:Cross} except number of iterations of a particular model, which the model determine dynamically. Finally, we applied simple majority vote ensemble \cite{Dietterich2000} on all 200 hard prediction on the test dataset. Table ~\ref{tab:finaltable} shows the testing scores from the final phase of the competition \footnote{\url{https://competitions.codalab.org/competitions/16283\#results}}.

\begin{table}
 \caption{Models accuracy on test dataset (Final standing)}
  \centering
  \begin{tabular}{ll}
    \toprule    
    Model     & Accuracy \\
    \midrule
    Ours (First-place team)      & 0.7200  \\
    Second-place team            & 0.7018  \\
    Third-place team             & 0.6909  \\
    Median competition scores    & 0.6436   \\ 
    Average competition scores   & 0.6181  \\
    \bottomrule
  \end{tabular}
  \label{tab:finaltable}
\end{table}

\section{Conclusion}
We introduced an ensemble learning approach to predict gender of shearwater
based on trajectory and associated metadata. The ensemble learning approach was able to model trajectories to output the predictions. One potential limitation of our approach is not considering cluster based information, which might be valuable to increase accuracy. As a future direction, we want to make cluster of grids from the trajectories and analyze the problem. One more interesting feature to see how 3D (latitude, longitude, and elevation) clustering might help increase accuracy. In aligned with this, it would be good to have date of the birds's trajectories, so that we might benefit from seasonal pattern of water level (elevation).

\bibliographystyle{unsrt}
\bibliography{references}

\end{document}